\documentclass[journal]{IEEEtran}
\usepackage{mathtools}

\usepackage{array}
\usepackage{caption}
\usepackage{multirow}
\usepackage{subcaption}
\usepackage{floatrow}
\usepackage{bbm}
\usepackage{amsfonts}
\usepackage{enumitem}
\usepackage[export]{adjustbox}
\usepackage[noadjust]{cite}

\newfloatcommand{capbtabbox}{table}[][\FBwidth]

\DeclareRobustCommand{\ghl}[1]{#1}
\DeclareRobustCommand{\chl}[1]{#1}
\newcommand{\gmhl}[1]{#1}
\newcommand{\cmhl}[1]{#1}

\usepackage{cite}

\ifCLASSINFOpdf
   \graphicspath{./Figures/}
\else
\fi
\begin{document}
%
\title{Reliable GNSS Localization Against Multiple Faults Using a Particle Filter Framework}
%
%
%

\author{Shubh~Gupta
        and~Grace~X.~Gao
\thanks{S. Gupta is with the Department
of Electrical Engineering, Stanford University,
CA, 94709, USA. e-mail: shubhgup@stanford.edu.}
\thanks{G. X. Gao is with the Department
of Aeronautics \& Astronautics, Stanford University,
CA, 94709, USA. e-mail: gracegao@stanford.edu.}}
\maketitle

\begin{abstract}
  For reliable operation on urban roads, navigation using the Global Navigation Satellite System (GNSS) requires both accurately estimating the positioning detail from GNSS pseudorange measurements and determining when the estimated position is safe to use, or available.
However, multiple GNSS measurements in urban environments contain biases, or faults, due to signal reflection and blockage from nearby buildings which are difficult to mitigate for estimating the position and availability. 
This paper proposes a novel particle filter-based framework that employs a Gaussian Mixture Model (GMM) likelihood of GNSS measurements to robustly estimate the position of a navigating vehicle under multiple measurement faults. 
Using the probability distribution tracked by the filter and the designed GMM likelihood, we measure the accuracy and the risk associated with localization and determine the availability of the navigation system at each time instant. 
Through experiments conducted on challenging simulated and real urban driving scenarios, we show that our method achieves small horizontal positioning errors compared to existing filter-based state estimation techniques when multiple GNSS measurements contain faults. 
Furthermore, we verify using several simulations that our method determines system availability with smaller probability of false alarms and integrity risk than the existing particle filter-based integrity monitoring approach.
\end{abstract}

\begin{IEEEkeywords}
  Particle filter, Integrity Monitoring, Global Navigation Satellite Systems, urban environment, mixture models, Sequential Monte Carlo, Expectation-Maximization.
\end{IEEEkeywords}

%
\IEEEpeerreviewmaketitle

\section{Introduction}
%
%
%
%

\begin{figure}[t!]
    \centering
    \includegraphics[width=0.8\linewidth]{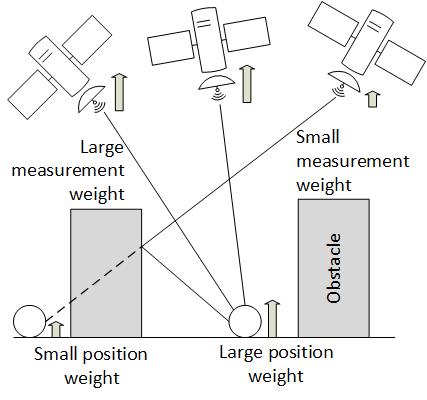}
    \caption{Our proposed algorithm for robust localization in the presence of several GNSS measurement faults associates weights both with the GNSS measurements and the different likely positions denoted by particles. We assign smaller weights to faulty GNSS measurements and to less likely positions while assigning large weights to the various positions supported by one or more non-faulty GNSS measurements.}
    \label{fig:title_fig}
\end{figure}

Global Navigation Satellite System (GNSS) in urban environments is often affected by limited satellite visibility, signal attenuation, non line-of-sight signals, and multipath effects~\cite{zhu_gnss_2018}. Such impairments to GNSS signals result in fewer measurements as compared to open-sky conditions as well as time-varying biases, or \textit{faults}, simultaneously in multiple measurements. Improper handling of these faults during localization may unknowingly result in large positioning errors, posing a significant risk to the safety of the navigating vehicle. Therefore, safe navigation in urban environments requires addressing two challenges: accurately estimating the position of the navigating vehicle from the faulty measurements and determining the estimated position's trustworthiness, or \textit{integrity}~\cite{zhu_gnss_2018}.

\subsection{Position Estimation under faulty GNSS Measurements}

For estimating the position of a ground-based vehicle, many existing approaches use GNSS measurements in tandem with odometry measurements from an inertial navigation system. Many of these approaches rely on state estimation by means of filters, such as Kalman filter and particle filter~\cite{morton_position_2020}. These filters track the probability distribution of the navigating vehicle's position across time, approximated by a Gaussian distribution in a Kalman filter and by a multimodal distribution represented as a set of samples, or particles, in a particle filter. However, the traditional filters are based on the assumption of Gaussian noise (or overbounds) in GNSS measurements, which is often violated in urban environments due to faults caused by multipath and non line-of-sight errors~\cite{zhu_gnss_2018}. 

Several methods to address non-Gaussian measurement errors in filtering have been proposed in the area of robust state estimation. Karlgaard~\cite{karlgaard_nonlinear_2015} integrated the Huber estimation technique with Kalman filter for outlier-robust state estimation. Pesonen~\cite{pesonen2007robust}, Medina \textit{et al.}~\cite{medina_robust_2016}, and Crespillo \textit{et al.}~\cite{crespillo_tightly_2018} developed robust estimation schemes for localization using fault-contaminated GNSS measurements. Lesouple~\cite{lesouple_multipath_2019} incorporated sparse estimation theory in mitigating GNSS measurement faults for localization. However, these techniques are primarily designed for scenarios where a large fraction of non-faulty measurements are present at each time instant, which is not necessarily the case for urban environment GNSS measurements.  

In the context of achieving robustness against GNSS measurement faults, several techniques have been developed under the collective name of Receiver Autonomous Integrity Monitoring (RAIM)~\cite{joerger_solution_2014-1}. RAIM algorithms mitigate measurement faults either by comparing state estimates obtained using different groups of measurements (solution-separation RAIM)~\cite{brown_self-contained_1988,lee_analysis_1986,joerger_solution_2014-1} or by iteratively excluding faulty measurements based on measurement residuals (residual-based RAIM)~\cite{parkinson_autonomous_1988,sturza_navigation_1988, joerger_sequential_2010, hongyu_optimal_2017}. Furthermore, several works have combined RAIM algorithms with filtering techniques for robust positioning. Grosch \textit{et al.}~\cite{grosch_snapshot_2017}, Hewitson \textit{et al.}~\cite{hewitson_extended_2010}, Leppakoski \textit{et al.}~\cite{leppakoski_raim_2006}, and Li \textit{et al.}~\cite{li_raim_2017} utilized residual-based RAIM algorithms to remove faulty GNSS measurements before updating the state estimate using KF. Boucher \textit{et al.}~\cite{boucher_non-linear_2004}, Ahn \textit{et al.}~\cite{ahn_gps_2011}, and Wang \textit{et al.}~\cite{wang_gps_2014,wang_fault_2018} constructed multiple filters associated with different groups of GNSS measurements, and used the logarithmic likelihood ratio between the distributions tracked by the filters to detect and remove faulty measurements. Pesonen~\cite{pesonen_framework_2011} proposed a Bayesian filtering framework that tracks indicators of multipath bias in each GNSS measurement along with the state. A limitation of these approaches is that the computation required for comparing state estimates from several groups of GNSS measurements increases combinatorially both with the number of measurements and the number of faults considered~\cite{joerger_solution_2014-1}. Furthermore, recent research has shown that poorly distributed GNSS measurement values can cause significant positioning errors in RAIM, exposing serious vulnerabilities in these approaches~\cite{sun_new_2019}.

In a recent work~\cite{gupta_particle_2019}, we developed an algorithm for robustly localizing a ground vehicle in challenging urban scenarios with several faulty GNSS measurements. To handle situations where a single position cannot be robustly estimated from the GNSS measurements, the algorithm employed a particle filter for tracking a multimodal distribution of the vehicle position. Furthermore, the algorithm equipped the particle filter with a Gaussian Mixture likelihood model of GNSS measurements to ensure that various position hypotheses supported by different groups of measurements are assigned large probability in the tracked distribution.

\subsection{Integrity Monitoring in GNSS-based Localization}

For assessing the trustworthiness of an estimated position of the vehicle, the concept of GNSS integrity has been developed in previous literature~\cite{pullen_sbas_nodate, zhu_gnss_2018}. Integrity is defined through a collection of parameters that together express the ability of a navigation system to provide timely warnings when the system output is unsafe to use~\cite{morton_position_2020, parkinson_global_1996}. Some parameters of interest for monitoring integrity are defined as follows
\begin{itemize}
    \item \textbf{Misleading Information Risk} is defined as the probability that the position error exceeds a specified maximum value, known as the alarm limit (AL).
    \item \textbf{Accuracy} represents a measure of the error in position output from the navigation system. 
    \item \textbf{Availability} of a navigation system determines whether the system output is safe to use at a given time instant.
    \item \textbf{Integrity Risk} is the probability of the event where the position error exceeds AL but the system is declared available by the integrity monitoring algorithm.
\end{itemize}
Different approaches for monitoring integrity have been proposed in literature. Solution-separation RAIM algorithms monitor integrity based on the agreement between the different state estimates obtained for different groups of measurements~\cite{brown_self-contained_1988,lee_analysis_1986,joerger_solution_2014-1}. Residual-based RAIM algorithms determine the integrity parameters from statistics derived from the measurement residuals for the obtained state estimate~\cite{parkinson_autonomous_1988,sturza_navigation_1988, joerger_sequential_2010, hongyu_optimal_2017}. \chl{Tanil \textit{et al.} }\cite{tanil_sequential_2018} utilized the innovation sequence of a Kalman filter to derive integrity parameters and determine the availability of the system. However, the approach is designed and evaluated for single satellite faults and does not necessarily generalize to scenarios where several measurements are affected by faults. However, these techniques have been designed for Kalman filter and cannot directly be applied to a particle filter-based framework. In Bayesian RAIM~\cite{pesonen_framework_2011}, the availability of the system is determined through statistics computed from the tracked probability distribution. The proposed technique is general and can be applied to a variety of filters, including the particle filter. However, the accuracy of the tracked probability distribution by a filter is limited in low-probability and tail regions necessary for monitoring integrity~\cite{panagiotakopoulos_extreme_2014}.

\subsection{Our Approach}


\ghl{In this paper, we build upon our prior work on a particle filter-based framework}~\cite{gupta_particle_2019} \ghl{that incorporates GNSS and odometry measurements both for estimating a position that is robust to faults in several GNSS measurements and for assessing the trustworthiness of the estimated position.} Unlike traditional particle filters used in GNSS-based navigation, our approach associates each GNSS measurement with a weight coefficient that is estimated along with particle filter weights. Our algorithm for estimating the measurement weights and the particle weights is based on the expectation-maximization algorithm~\cite{vila_expectation-maximization_2012}. At each time instant, our algorithm mitigates several faults at once by reducing the weights assigned to both the faulty GNSS measurements and the particles corresponding to unlikely positions (Fig.~\ref{fig:title_fig}). Using the measurements and the estimated weights, we then evaluate measures of misleading information risk, accuracy and determine the availability of the localization output.    


To reduce the effect of faulty GNSS measurements on the particle weights, we model the likelihood of GNSS measurements within the particle filter as a Gaussian mixture model (GMM)~\cite{geary_mixture_1989} with the measurement weights as the weight coefficients.
The GMM likelihood is characterized by a weighted sum of multiple probability distribution components totaling the number of available measurements at the time instant. 
Each component in our GMM represents the conditional probability distribution of observing a single GNSS measurement from a position. In scenarios where determining a single best position from the measurements is difficult, our designed GMM likelihood assigns high probability to several positions that are likely under different groups of GNSS measurements.

\begin{figure*}[t!]
        \centering
        \includegraphics[width=\linewidth]{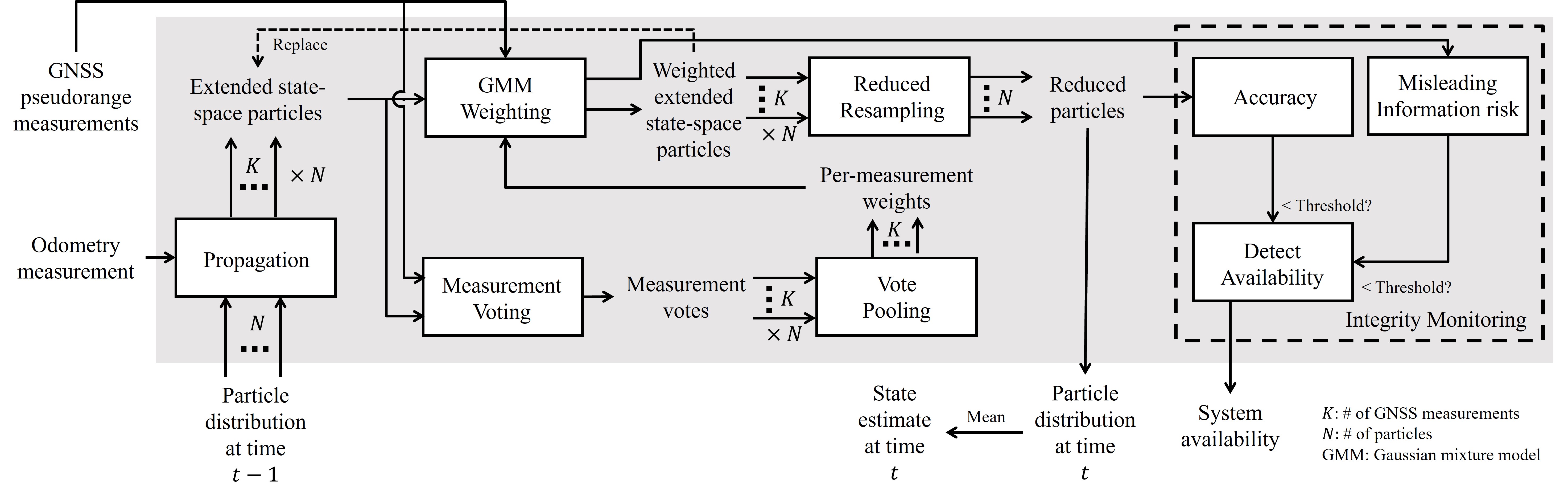}
        \caption{\ghl{Architecture of the proposed framework. The framework incorporates GNSS and odometry measurements to infer the probability distribution of the vehicle state as a set of weighted particles in a particle filter. Robustness to faults in GNSS measurements is achieved using the GMM Weighting, Measurement Voting and Vote Pooling steps which are based on expectation-maximization~\cite{vila_expectation-maximization_2012}. From the estimated probability distribution and the GNSS measurements, the framework derives measures of Accuracy and Misleading Information risk to determine the availability of the system at each time instant.}}
        \label{fig:arch}
    \end{figure*}

We next describe the key contributions of this work:
\begin{enumerate}
    \item We develop a novel particle filter-based localization framework that mitigates faults in several measurements by estimating the particle weights together with weight coefficients associated with GNSS measurements. Our approach for estimating the measurement and particle weights integrates the weighting step in the particle filter algorithm~\cite{doucet_tutorial_2009} with the expectation-maximization algorithm~\cite{dempster_maximum_1977}. 
    
    \item We design an approach for determining the navigation system availability in the presence of faults in several GNSS measurements. Our designed approach considers both the variance in the particle filter probability distribution like~\cite{pesonen_framework_2011} and the  disagreement between residuals computed from different GNSS measurements.
    
    \item \ghl{We have validated our approach on real-world and simulated driving scenarios under various scenarios to verify the statistical performance. We have compared the localization performance of our proposed framework against existing Kalman filter and particle filter-based baselines, and the integrity monitoring performance against the existing particle filter-based method.}
        
\end{enumerate}

\ghl{Our approach provides several advantages over prior approaches for fault-robust localization and integrity monitoring.}
While inspired by existing residual-based RAIM algorithms which mitigate faults using residuals computed from an initial position estimate, our approach utilizes the tracked probability distribution of vehicle position to compute the residuals and mitigate faults. This enhances the robustness of our approach in scenarios with multiple faults since several different position hypotheses are considered. \ghl{Unlike the existing solution-separation RAIM and filter-based approaches that exhibit a combinatorial growth in computation with increasing number measurements and measurement faults, our approach scales linearly in computation with the number of measurements. 
This is due to our weighting scheme that simultaneously assigns weights to GNSS measurements and particles instead of individually considering several subsets of the measurements. Since our approach tracks several position hypotheses for mitigating measurement faults, it can track the position probability distribution in scenarios with a large fraction of faulty GNSS measurements, unlike the existing robust state estimation approaches. Unlike existing particle filter-based approaches that rely on a known map of the environment~\cite{hsu_nlos_2015,miura_gps_2013,hsu_evaluation_2015} or a known transition model of faults~\cite{pesonen_framework_2011}, our approach requires no assumptions about existing environment information for mitigating faults. 
Furthermore, our measure of the misleading information risk for integrity monitoring is derived through our designed GMM likelihood of measurements, which accounts for the multi-modal and heavy-tailed nature of the position probability distribution, unlike the purely filter-based approach presented in }~\cite{pesonen_framework_2011}.

The rest of the paper is organized as follows: Section II discusses the proposed algorithm in detail. Section III details the experiments and comparisons with existing techniques on both simulated and real-world data. Section IV summarizes our work and provides concluding remarks.

\section{Localization and Integrity Monitoring Framework}

    The different components of our method are outlined in Fig.~\ref{fig:arch}
\begin{enumerate}
    \item \chl{The \textbf{Propagation step} receives as input the previous time particle distribution of receiver state and the odometry measurement and generates as output propagated particles in the \textit{extended state-space}, which are associated with different GNSS measurements. 
    The extended state-space consists of the receiver state along with an integer-valued selection of a GNSS measurement associated with the particle.}
    \item \chl{The \textbf{Iterative Weighting step} receives as input the extended state-space particles and the GNSS measurements and infers the particle weights and GMM weight coefficients in three steps: 
    First, the \textbf{Measurement Voting step} computes a probabilistic confidence, referred to as vote, in the association between the extended state-space particle and the chosen GNSS measurement. 
    Next, the \textbf{Vote Pooling step} combines the votes to obtain an overall confidence in each GNSS measurement, referred to as measurement weights.
    Finally, the \textbf{GMM Weighting step} updates the particle weights using the a GMM constructed from the measurement weights as the measurement likelihood function. The process is then repeated for a fixed number of iterations, or till convergence.}
    \item \chl{The \textbf{Reduced Resampling step} receives as input the extended state-space particles and their computed weights and generates a new set of particles in the receiver state space.}
    \item \chl{For IM, we compute measures of \textbf{Accuracy} and \textbf{Misleading Information Risk} using the particles and the GMM likelihood. 
    Then, we determine the \textbf{System Availability} by comparing the obtained accuracy and misleading information risk against prespecified thresholds.}
\end{enumerate}
We next describe our problem setup, followed by details about the individual components of our method. Finally, we end the section with the run-time analysis of our approach.

    \subsection{Problem Setup}
    We consider a vehicle navigating in an urban environment using GNSS and odometry measurements. We denote the true state vector of the vehicle at time $t$ as $x_t$, and the overall trajectory as $\{x_t\}_{t=1}^T$, where $T$ denotes the total duration of the vehicle's trajectory. For generality, we do not restrict $x_t$ to have a specific form and instead assume that it consists of various quantities such as the vehicle's position coordinates, \ghl{heading angles}, receiver clock errors, linear velocities, and yaw rates. The $k$th GNSS pseudorange measurement acquired at time $t$ is denoted by the scalar value $\rho_t^{(k)}$. At time $t$, the vehicle acquires a set of $K_t$ GNSS pseudorange measurements $M_t = \{\rho_t^{(k)}\}_{k=1}^{K_t}$ and odometry measurement $u_t$. The probability distribution $\pi_t$ of the vehicle state is approximated by a particle filter as
    \begin{equation}
        \pi_t = P(x_t | x_{t-1}, M_t, u_t) \cmhl{\approx \hat{\pi}_t} = \sum_{i=1}^N w_t^{(i)}\delta(x=x_t^{(i)}),     
    \end{equation} 
    where \chl{$\hat{\pi}_t$ denotes the approximate state probability distribution represented using particles}; $x_t^{(i)}$ denotes the state of the $i$th particle; $w_t^{(i)}$ denotes the weight of the $i$th particle such that $\sum_{i=1}^Nw_t^{(i)}=1$; $\delta(x=x_t)$ denotes the Dirac delta function~\cite{dirac_principles_1981} that assumes the value of zero everywhere except $x_t$; and $N$ is the total number of particles.   

    To design our particle filter-based localization algorithm, we assume that the state of the vehicle follows the Markov property, i.e. given the present state of the vehicle, the future states are independent of the past states. By the Markov property among $(x_1,\ldots,x_t)$, the conditional probabilities $P(x_t | x_{t-1}, M_t, u_t)$ and $P(x_t | x_1,\ldots,x_{t-1}, M_1, \ldots, M_t, u_t)$ are equivalent. Following the methodology in filtering algorithms~\cite{chen_bayesian_2003} and applying Bayes' theorem with independence assumptions on $M_t, u_t$, the probability distribution is factored as
    \begin{align}    
            P(x_t | x_{t-1}, M_t, u_t)  &\propto P(M_t | x_t, x_{t-1}, u_t) \cdot P(x_t | x_{t-1}, u_t) \nonumber\\
            &\propto \underbrace{P(M_t | x_t)}_{L_t(x_t)} \cdot \underbrace{P(x_t | x_{t-1}, u_t)}_{\tilde{\pi}_t}, 
    \end{align}
    where $L_t(x_t)$ can be interpreted as the likelihood of receiving $M_t$ at time $t$ from $x_t$; and $\tilde{\pi}_t$ denotes the probability distribution at time $t$ predicted using $\hat{\pi}_{t-1}$ and $u_t$.
    
    In our approach, we consider the joint tasks of state estimation and fault mitigation using an extended state-space $(x,\chi)$ of particles, where $\chi = k \in \{1, \ldots, K_t\}$ denotes an integer corresponding to the $k$th GNSS measurement in $M_t$. The value of $\chi$ is assigned to each extended state-space particle at the time of its creation in the propagation step, and remains fixed until the resampling step. The extended state-space particles are used in the iterative weighting step to determine the weight coefficients in the GMM likelihood as well as to compute the particle weights.

    \subsection{Propagation Step}
    First, we generate uniformly weighted $K_t$ copies of each particle $x_{t-1}^{(i)}$, each corresponding to a different value of $\chi = \{1, \ldots, K_t\}$. We then propagate these extended state-space particles \ghl{$(x)_{t-1}^{(i, \chi)}$} via the state dynamics
    \begin{align}
        \gmhl{x_t^{(i, \chi)}} &\gmhl{= f(x_{t-1}^{(i, \chi)}, u_t) + \epsilon^{(i, \chi)}_f \quad \forall \  \chi=\{1,\ldots,K_t\},} \\ 
        \tilde{\pi}^{(i)}_{t} &= \sum_{k=1}^{K_t} K_t^{-1} \delta(x=\tilde{x}_t^{(i, \chi=k)}),
        \\
        \tilde{\pi}_t &\approx \sum_{i=1}^N w_{t-1}^{(i)}\tilde{\pi}^{(i)}_{t} ,
    \end{align}
    where \ghl{$f:\mathcal{X} \times \mathcal{U} \to \mathcal{X}$ is the function to propagate state $x \in \mathcal{X}$ using odometry $u \in \mathcal{U}$ based on system dynamics}; and \chl{$\epsilon^{(i, \chi)}_f \sim \mathcal{N}(0, \sigma^2_f)$ denotes the stochastic term in propagation of each particle}.
    
    Next, we compute the measurement likelihood model and update the weights associated with each particle in the weighting step.

    \subsection{Iterative Weighting Step}
    In conventional filtering schemes for GNSS, the measurement likelihood model is designed with the assumption that all the measurements are independent of each other and model the probability $P(\rho_t^{(k)}|x_t)$ associated with the $k$th measurement at time $t$ for $x_t$ as a Gaussian distribution~\cite{chen_bayesian_2003,joerger_kalman_2012,pesonen_framework_2011,yu_research_2013}. However, this likelihood model inherently assumes a unimodal probability distribution of $x_t$, and therefore, has a tendency to over simplify \ghl{the measurement errors} when faults occur in multiple GNSS measurements. 
    
    Hence, we model the likelihood instead as a Gaussian mixture model (GMM) which has additive contributions of each measurement towards the likelihood, such that faults in any small subset of measurements does not have a dominating effect on the overall likelihood. Each measurement component in the GMM likelihood is associated with a weight coefficient which controls the impact of that component on the overall likelihood. The mixture model based likelihood is expressed as
    \begin{equation}
        L_t(x_t) = \sum_{k=1}^{K_t}\gamma_k \mathcal{N}\left(\rho_t^{(k)}\mid\hat{\rho}_{x_t}^k, \left(\sigma_{x_t}^k\right)^2\right) \  \text{s.t.} \  \sum_{k=1}^{K_t}\gamma_k = 1,
    \end{equation}
    where $\hat{\rho}_{x_t}^{k}$ is the expected value of the $k$th pseudorange measurement from $x_t$ assuming known satellite position; $\sigma_{x_t}^{k}$ is the standard deviation obtained using the carrier-to-noise ratio, satellite geometry or is empirically computed~\cite{zhao_improved_2011}; and \ghl{$\gamma_k$ is the measurement weight associated with $k$th measurement that scales the individual probability distribution components to enforce a unit integral for the probability distribution.} A single measurement component for the $k$th measurement in the GMM assigns similar probabilities to all $x_t$ which generate similar values of $\hat{\rho}_{x_t}^k$. 
    Therefore, the GMM induces a multi-modal probability distribution over the state, with peaks at positions supported by different subsets of component measurements. The values of the measurement weights decide the relative confidence between the modes, i.e., for the same values of standard deviation, the mode of the component with higher measurement weight has a higher probability value associated with it~\cite{geary_mixture_1989}.  
    
    Using the proposed GMM likelihood, we jointly determine the values for $\{\gamma_k\}_{k=1}^{K_t}$ and $\{w^{(i)}_t\}_{i=1}^{N}$ using an iterative approach based on the EM algorithm for GMM~\cite{vila_expectation-maximization_2012}. Our approach consists of three steps, namely Measurement Voting, Vote Pooling and GMM Weighting.  

    \begin{enumerate}
        \item \ghl{\textbf{Measurement Voting} 
        Assuming no prior knowledge about measurement faults,} we uniformly initialize the value of $\gamma_k$ with $K_t^{-1}$ for all values of $k \in \{1,\ldots,K_t\}$. The initial value of $\{w^{(i, \chi=k)}_t\}_{i=1}^N$ is set using the previous time instant weights as
       \begin{equation}
           w^{(i, \chi=k)}_t = K_t^{-1} \cdot  w^{(i)}_{t-1}. 
       \end{equation} 
       
       where $i \in \{1,\ldots,N\}$.
       Since $\{w^{(i)}_t\}_{i=1}^N$ and measurement weights $\{\gamma_k\}_{k=1}^{K_t}$ are \ghl{interdependent}, we cannot obtain a closed form expression for their updated values. Therefore, we introduce a set of random variables $\{V^k\}_{k=1}^{K_t}$, referred to as \textit{votes}, to indirectly compute the updates for both $\{w^{(i)}_t\}_{i=1}^N$ and $\{\gamma_k\}_{k=1}^{K_t}$. The random variable $V^k$ denotes the confidence associated with the $k$th measurement depending on a given value of the state $x_t$, and differs from $\gamma_k$ due to its dependence on $x_t$. This dependence is exercised by computing the normalized residual $r^k_i$ for each $x_t^{(i, \chi)}$ in $\tilde{\pi}_t$
       \begin{equation}
           r^k_i = (\sigma_{x^{(i, \chi=k)}_t}^k)^{-1}(\rho_t^{(k)} - \hat{\rho}_{x^{(i, \chi=k)}_t}^k).
       \end{equation}
       Using the initial values of $\{w^{(i)}_t\}_{i=1}^N$ and $\{\gamma_k\}_{k=1}^{K_t}$ and the computed residuals $\{r^k_i\}_{i=1, k=1}^{N, K_t}$, we then infer the probability distribution of the random variables $\{V^k\}_{k=1}^{K_t}$ in the expectation-step of the EM algorithm~\cite{vila_expectation-maximization_2012}. We model the probability distribution of $\{V^k\}_{k=1}^{K_t}$ as a weighted set of votes $\{v^k_i\}_{i=1,k=1}^{N,K_t}$
       \begin{equation}
           v^k_i = P_{\mathcal{N}^2(0, 1)}\left(r^{k}_i\right)^2,
       \end{equation}
       where $P_{\mathcal{N}^2(0, 1)}(\cdot)$ denotes the probability density function of the square of standard Gaussian distribution~\cite{simon_probability_2007}. \ghl{The squared standard Gaussian distribution assigns smaller probability value to large residuals than the Gaussian distribution.} This is in line with RAIM algorithms that use the chi-squared distribution for detecting GNSS faults~\cite{parkinson_autonomous_1988}.
    \item \ghl{\textbf{Vote pooling}
    Next, we normalize and pool the votes cast by each particle in the maximization-step to compute $\gamma_k$ for each measurement.} We write an expression for the total empirical probability \ghl{$\pi_t^{tot}$} at time $t$ of measurements $M_t$ using the votes $\{v^k_i\}_{i=1, k=1}^{N, K_t}$ and measurement weights $\{\gamma_k\}_{k=1}^{K_t}$ as
    \begin{equation}
        \gmhl{\pi_t^{tot}} = \sum_{i=1}^N \sum_{k=1}^{K_t} \gamma_k w^{(i, \chi=k)}_t v^k_i.
    \end{equation}
    We maximize $\pi_t^{tot}$ with respect to $\{\gamma_k\}_{k=1}^{K_t}$ and the constraint $\sum_{k=1}^{K_t}\gamma_k = 1$ to obtain an update rule for vote pooling
    \begin{equation}
        \gamma_k = \frac{\sum_{i=1}^N w^{(i, \chi=k)}_t v^k_i}{\sum_{i=1}^N \sum_{k'=1}^{K_t} w^{(i, \chi=k)}_t v^{k'}_i}.
    \end{equation}
    This update rule can be seen as assigning an empirical probability to each $\gamma_k$ according to the weighted votes.

    \item \textbf{GMM weighting}
    Using the computed measurement weights $\{\gamma_k\}_{k=1}^{K_t}$, we then update the weights of each extended state-space particle. We compute the updated particle weights in the logarithmic scale, since implementing the particle filter on a machine with finite precision of storing floating point numbers may lead to numerical instability~\cite{gentner_corrigendum_2018}. However, the GMM log-likelihood contains additive terms inside the logarithm and therefore cannot be readily distributed. Hence, we consider an extended state-space that includes the measurement association variable $\chi$ described in Section II.A. \ghl{The measurement likelihood} conditioned on $\chi$ can now be equivalently written as a Categorical distribution likelihood~\cite{bishop_pattern_2006}
    \begin{flalign}
        &L_t(x_t) = P(M_t | x_t, \chi) = \prod_{k=1}^{K_t} \left(\gamma_k \mathcal{N}_{x_t}(\rho^{(k)}_t)\right)^{\mathbb{I}[\chi=k]}_{\textstyle,}&
        \raisetag{20pt}
    \end{flalign}
    where $\mathbb{I}[\cdot]$ denotes an indicator function which is 1 if the condition in its argument is true and zero otherwise; $\mathcal{N}_{x_t}(\rho^{(k)}_t)$ is shorthand notation for $\mathcal{N}\left(\rho^{(k)}_t \mid \hat{\rho}_{x_t}^k, \left(\sigma_{x_t}^k\right)^2\right)$; and $\sum_{k=1}^{K_t} \gamma_k = 1$. The log-likelihood from the above expression is derived as
    \begin{flalign}
        &\log L_t(x_t) = \sum_{k=1}^{K_t} \mathbb{I}[\chi=k] \left( \log\gamma_k + \log \mathcal{N}_{x_t}(\rho^{(k)}_t) \right),&
    \end{flalign}
    We use the log-likelihood derived above to compute the new weights $\{w^{(i, \chi=k)}_t\}_{i=1, k=1}^{N, K_t}$ for each particle in a stable manner
    \begin{align}
        l^{(i)}_k = \log & L_t x^{(i, \chi=k)}_t - \max_{i, k} \left( \log L_t x^{(i, \chi=k)}_t \right), 
        \\
        w^{(i, \chi=k)}_t &= \frac{\exp\left(l^{(i)}_k\right)}{\sum_{i=1}^N\sum_{k=1}^{K_t} \exp\left(l^{(i)}_k\right)}.        
    \end{align}
    
    \end{enumerate}

    \subsection{Reduced Resampling Step}
     To obtain the updated particle distribution $\{x_t^{(i)}\}_{i=1}^{N} = \hat{\pi}_t$, we redistribute the weighted extended state-space particles via the sequential importance resampling (SIR) procedure~\cite{gustafsson_particle_2002}. Additionally, we reduce the $N \times K_t$ particles to the original number of $N$ particles with equal weights
    \begin{equation}
        \{x_t^{(i)}\}_{i=1}^{N} \leftarrow \text{SIR}\left(\{(x)_t^{(i, \chi=k)}, w_t^{(i, \chi=k)}\}_{i=1, k=1}^{N, K_t}\right),
    \end{equation}
    where $\text{SIR}(\cdot)$ is the SIR procedure~\cite{gustafsson_particle_2002} that resamples from the categorical distribution of weighted extended space particles.
    The mean state estimate $\hat{x}_t$ denotes the state solution from the algorithm, and is computed as $\hat{x}_t = \sum_{i=1}^{N} w_t^{(i)}x_t^{(i)}$.
     
\subsection{Integrity Monitoring}
    \chl{A key feature of our integrity monitor is that we consider a non-Gaussian probability distribution of the receiver state represented through particles. 
    Our integrity monitor is based on two fundamental assumptions: 
    First, sufficient redundancy exists in positioning information across the combination of measurements from different satellites and multiple time epochs. 
    Second, the positions which are likely under faulty measurements have lower correlation with the filter dynamics across time than the positions which are likely under non-faulty measurements. }

    \ghl{We develop a Bayesian framework for integrity monitoring inspired by}~\cite{pesonen_framework_2011}.
    \ghl{At each time instant, the integrity monitor calculates measures of misleading information risk $P_{MIR}$ (referred to as integrity risk in~\cite{pesonen_framework_2011}) and accuracy $r_{A}$, which are derived later in the section. The integrity monitor then compares the misleading information risk $P_{MIR}$ and accuracy $r_{A}$ against prespecified reference values to detect whether navigation system output is safe to use.} \chl{Following the Stanford-ESA integrity diagram }\cite{tossaint_stanford_2007}, \chl{we refer to such events as hazardous operations. If hazardous operations are detected, the integrity monitor declares the system as unavailable.}      

    \paragraph{Misleading Information Risk}
     In~\cite{pesonen_framework_2011}, the process of computing $P_{MIR}$ relies on the state probability distribution tracked by several Kalman or particle filters. The filters are characterized by considering multipath errors in different measurements, thus tracking different position probability distributions. $P_{MIR}$ is then measured by integrating the probability over all positions that lie outside the specified AL.  However, this approach for estimating $P_{MIR}$ is often inaccurate, since both the Kalman filter and the particle filter algorithms are designed to estimate the probability distribution in the vicinity of the most probable states and not the tail-ends~\cite{van_der_merwe_unscented_2001}. Therefore, we derive an alternative approach to estimate $P_{MIR}$ directly from our GMM measurement likelihood
    \begin{flalign*}
        \cmhl{P_{MIR}} &\cmhl{= P(\{x_t \not\in \Omega_I\} \mid M_t, u_t, \hat{\pi}_{t-1})}  \\
        \cmhl{= 1 - &\frac{P(\{x_t \in \Omega_I\} \mid u_t, \hat{\pi}_{t-1})P(M_t \mid \{x_t \in \Omega_I\}, u_t, \hat{\pi}_{t-1}) }{P(M_t \mid u_t, \hat{\pi}_{t-1})}}  \\
        \cmhl{\approx \ \  &\hat{P}_{MIR} = 1 - \frac{P(\{x_t \in \Omega_I\} \mid u_t, \hat{\pi}_{t-1})}{P(M_t \mid u_t, \hat{\pi}_{t-1})}
        \\
        & \int_{x_t \in \Omega_I} |\Omega_I|^{-1}\sum^{K_t}_{k=1}\gamma_k\mathcal{N}\left(\rho_t^{(k)} \mid \hat{\rho}_{x_t}^k, \left(\sigma_{x_t}^k \right)^2 \right) dx_t} \addtocounter{equation}{1}\tag{\theequation}
    \end{flalign*}
    where $\Omega_{I}$ denotes the set of positions that lie within AL about the mean state estimate $\hat{x}_t$ and \chl{$|\Omega_I|$ denotes its total area. For simplicity, we approximate the conditional distribution ${P(x_t \mid \{x_t \in \Omega_I\}, u_t, \hat{\pi}_{t-1})}$ as a uniform distribution $|\Omega_I|^{-1} \ \forall x_t \in \Omega_I$. We empirically verify that this approximation works well in practice for determining the system availability.}

    For computational efficiency, we approximate the above integral using cubature techniques for two-dimensional disks described in~\cite{lether_generalized_1971}. \chl{The term $P(\{x_t \in \Omega_I\} \mid u_t, \hat{\pi}_{t-1})$ is computed by adding the weights of particles that lie inside AL based on the propagated distribution $\tilde{\pi}_t$.}  
    
    \paragraph{Accuracy} Bayesian RAIM~\cite{pesonen_framework_2011} defines $r_A$ as the radius of the smallest disk about $\hat{x}_t$ that contains the vehicle state with a specified probability $\alpha$. Unlike $P_{MIR}$, accuracy is determined primarily by the probability mass near the state estimate and not the tail-ends. To compute accuracy, we first approximate the particle distribution by a Gaussian distribution with the mean parameter as $\hat{x}_t$ and covariance $\mathbf{C}$ computed as a weighted estimate from samples as
    \begin{equation}
        \mathbf{C} = \frac{1}{1-\sum_{i=1}^N \left(w^{(i)}_t\right)^2}\sum_{i=1}^N w^{(i)}_t (x^{(i)}_t - \hat{x}_t)(x^{(i)}_t - \hat{x}_t)^\top.
    \end{equation}
    Next, we estimate the accuracy $r_{A}$ using the inverse cumulative probability distribution of the Gaussian distribution as
    \begin{equation}
        r_{A} \cmhl{\approx \hat{r}_{A}} = \max_{i = 1,2} \sqrt{\mathbf{C}_{i,i}} \Phi^{-1}(\alpha)
    \end{equation}
    where \chl{$\hat{r}_{A}$ is the estimated accuracy}; $\Phi^{-1}(\cdot)$ denotes the inverse cumulative probability function for standard Gaussian distribution.
    \chl{A smaller value of $r_{A}$ computed by the expression implies higher accuracy in positioning.}

    \paragraph{Availability} We compare the estimated values of $\hat{P}_{MIR}$ and $\hat{r}_A$ against their respective thresholds $P_{MIR}^0$ and $r^0_{A}$ specified by the user
    \begin{equation}
        \cmhl{(\hat{P}_{MIR} \le P_{MIR}^0)\quad \text{and}\quad (\hat{r}_{A} \le r_{A}^0).}
    \end{equation}
    If any of the above constraints are violated, the integrity monitor \ghl{declares the system unavailable}.

    \subsection{Computation Requirement}
    We analyze the computational requirement of our algorithm and compare it with the requirement of the existing least squares, robust state estimation, residual-based RAIM and particle filter-based approaches. The required computation in the existing least-squares and robust state estimation approaches (both filter based and non-filter based) is proportional to the number of available measurements $k$, since each measurement is only seen once to compute the position. For residual-based RAIM algorithms that remove faults iteratively~\cite{parkinson_autonomous_1988,sturza_navigation_1988, joerger_sequential_2010, hongyu_optimal_2017}, the maximum required computation grows proportionally to $mk$ since it depends both on the number of iterations (maximum $m$) and on the computation required per iteration (proportional to $k$). For the existing particle filter-based methods that employ several filters~\cite{boucher_non-linear_2004, wang_gps_2014,ahn_gps_2011,wang_fault_2018}, the computation depends on the number of filters, which scales proportionally to $k^m$.      
    
    Instead, our approach grows linearly in computation with the number of available measurements. For $n$ particles and $k$ available measurements, each component in our framework has a maximum computational requirement proportional to $nk$ irrespective of the number of faults present. Hence, our approach exhibits smaller computational requirements than existing particle filter-based methods, and similar requirements to residual-based RAIM, least-squares and robust state estimation algorithms with respect to the number of GNSS measurements.
    
    
    \section{Experimentation Results}
We evaluated the framework for both the localization and integrity monitoring performance and on simulated and real-world driving scenarios. For the experiments on simulated data, we validated our algorithm across multiple trajectories with various simulated noise profiles that imitate GNSS measurements in urban environments.  

For our first set of experiments on evaluating the localization performance under GNSS measurement faults, we consider two baselines:

\begin{enumerate}[label=\alph*)]
    \item Joint state-space particle filter (J-PF), which is based on the particle filter algorithm proposed by Pesonen~\cite{pesonen_framework_2011} \ghl{which considers both the vehicle navigation parameters and measurement fault vector (signal quality parameters) in the particle filter state-space.} \chl{We refer to the combined state-space comprising of all the parameters as the \emph{joint state-space}.} Since J-PF independently considers both the measurement faults and the state variables, it can also be interpreted as a bank of particle filters~\cite{wang_fault_2014, wang_fault_2018}. Since computational requirement of J-PF is combinatorial in the number of considered measurement faults, we limit the algorithm to consider at most two faults. Furthermore, we assume a uniformly distributed fault transition probability in the particle filter. 
    \item Kalman filter RAIM (KF-RAIM), which \chl{combines the residual-based RAIM algorithm}~\cite{parkinson_autonomous_1988} \chl{with Kalman filter for fault-robust sequential state estimation. Similar to the fault exclusion strategy employed in}~\cite{leppakoski_raim_2006} and \cite{gunning_integrity_2019}, \chl{KF-RAIM detects and removes measurement faults iteratively by applying a global and a local test using statistics derived from normalized pseudorange measurement residuals. The local test is repeated to remove faulty measurements until the global test succeeds and declares the estimated position safe to use.} 
\end{enumerate}

For our second set of experiments for evaluating the performance in determining the system availability, we compare our approach against Bayesian RAIM~\cite{pesonen_framework_2011}, which determines the system availability through measures of misleading information risk and accuracy computed by integrating the J-PF probability distribution.


\ghl{All the filters are initialized at the ground truth initial position with a specified standard deviation $\sigma_{init}$. Note that for our approach, we only used a single iteration in the weighting step in our experiments on simulated data, since we observed that a single iteration suffices to provide us good performance at low computation cost.}

\paragraph{Localization Performance Metrics}
\chl{
For evaluating the positioning performance of our approach, we compute the metrics of horizontal root mean square error (RMSE) as well as the percentage of estimates that deviate more that 15 m from the ground truth ($\%{>}15$ m). We compute these metrics as
}
\begin{equation}
    \cmhl{R M S E = \sqrt { \frac { 1 } { T } \sum _ { t = 1 } ^ { T } \| \hat { x } _ { t } - x _ { t } ^ { * } \|_{\text{pos}} ^ { 2 } },}
\end{equation}
\begin{equation}
    \cmhl{\%{>}15\text{ m} = \frac { N ( \| \hat { x } _ { t } - x _ { t } ^ { * } \| > 15 ) } { T },}
\end{equation}
where $T$ denotes the total number of timesteps for which the algorithm is run; $x^*_t$ denotes the ground truth state of the receiver; $\|x\|_{\text{pos}}$ denotes the Euclidean norm of the horizontal position coordinates in state $x$; and $N(I)$ denotes the number of occurrences of event $I$.

\paragraph{Integrity Monitoring Performance Metrics}
\ghl{
We evaluate the performance of our integrity monitor on its capability to declare the navigation system unavailable when the navigation system is under hazardous operations. In our experiments, we refer to hazardous operations as the event when the positioning error in the navigation system exceeds an alarm limit, which we set to $15$ m. For evaluation, our metrics consist of estimated probability of false alarm $\hat{P}(FA)$ and integrity risk $\hat{P}(IR)$ computed as}

\begin{equation}
    \hat{P}(FA) = \frac{N(\neg I_{av}\bigcap \neg I_{ho})}{T},
\end{equation}
\begin{equation}
    \hat{P}(IR) = \frac{N(I_{av}\bigcap I_{ho})}{T},
\end{equation}
where $I_{av} (\neg I_{av})$ denotes the event when the navigation system is declared available (unavailable) by the IM algorithm and $I_{ho} (\neg I_{ho})$ denotes the event of hazardous operations (normal operations).

The rest of the section is organized as follows: (A) evaluation of the localization performance on simulated data and comparison of our approach with KF-RAIM and J-PF approaches, (B) evaluation of our approach on real-world data, (C) comparison of our integrity monitoring algorithm with Bayesian RAIM~\cite{pesonen_framework_2011} on simulated scenarios, (D) analysis of the computation time of our algorithm and comparison in terms of computation with KF-RAIM and J-PF approaches, and (E) discussion of the results.


\subsection{Evaluation on simulated scenarios}
\ghl{We evaluated our approach on multiple simulated driving scenarios with varying noise profiles in GNSS and odometry measurements. 
Our choice of simulations is motivated by two factors: accurate ground truth information is available and multiple measurement sequences with different noise values can be obtained. 
Our simulation environment consists of a vehicle moving on the horizontal surface according to 50 different randomly generated trajectories approximately of length $4000$ m each}.
\ghl{The vehicle obtains noisy GNSS ranging measurements from multiple simulated satellites as well as noisy odometry measurements. 
The simulated satellites move with a constant velocity of $1000$ m/s along well-separated paths in random directions at fixed heights of $2 \times 10^7$ m from the horizontal surface.  
The odometry consists of vehicle speed measurements, and an accurate heading direction of the vehicle is assumed to be known (eg. from a high accuracy magnetometer). 
A driving scenario lasts for $400$ s with measurements acquired at the rate of $1$ Hz. 
In each state estimation algorithm, we track the 2D state $(x,y)$ of the vehicle. 
We limit ourselves to the simple 2D case in simulation in view of the computational overhead from requiring significantly more particles in larger state-spaces.
The parameters used for the simulation and filter are given in Table}~\ref{tab:param}. 
 
\begin{table}[t!]
    \centering
    \begin{tabular}{@{}ll@{}}
        \hline
        Parameter & Value \\
        \hline
        \# of satellites & $5$-$10$\\
        \ghl{GNSS ranging noise $\sigma_{GNSS}$} & $5$-$10$ m \\
        GNSS ranging bias error magnitude & $50$-$200 $ m\\
        Maximum \# of faulty measurements & $1$-$6$\\
        GNSS fault change probability & $0.2$ \\ 
        Vehicle speed & $10 $ m/s \\
        \ghl{Odometry noise $\sigma_u$} & $5 $ m/s \\
        \# of particles & $500$ \\
        \ghl{Filter initialization $\sigma_{init}$} & $5 $ m \\
        \ghl{Filter measurement model noise $\{\sigma^k\}_{k=1}^K$} & $5$ m \\
        \ghl{Filter propagation model noise $\sigma_f$} & $5 $ m \\
        \chl{Alarm limit $AL$} & $15$-$20$ m \\
        \chl{Accuracy probability requirement $\alpha$} & $0.5$ \\
        \hline
    \end{tabular}
    \caption{\ghl{Experimental parameters for simulated scenario}}%
     \label{tab:param}%
\end{table}

We induce two kinds of noise profiles in our pseudorange measurements from the simulation - a zero mean Gaussian random noise on all measurements and a random bias noise on a subset of measurements. 
The subset of measurements containing the bias noise is initialized randomly, and has a small probability of changing to a different random subset at every subsequent time instant. \ghl{The number of affected measurements is selected randomly between zero and a selected maximum number of faults at every change.} The zero mean Gaussian noise is applied to measurements such that the variance is double the normal value for biased measurements.
Using our noise model, we simulate the effects of non line-of-sight signals and multipath on pseudorange measurements in urban GNSS navigation.  


\paragraph{Localization performance}
\begin{table}[t!]
    \centering
    \begin{tabular}{@{}lllllll@{}}
    \hline
    && \multicolumn{2}{c}{Few faults} && \multicolumn{2}{c}{Many faults} \\
    \cline{3-4} \cline{6-7} 
    && $(5, 1)$ & $(5, 2)$ && $(7, 4)$ & $(10, 6)$\\ 
    \hline
    \multirow{2}{*}{KF-RAIM} & RMSE(m) & $7.6$ & $21.5$ && $34.3$ & $34.4$\\
    &$\%{>}15$ m &  $5.9$ & $68.7$ && $87.4$ & $91.1$\\
    \multirow{2}{*}{J-PF} & RMSE(m) & $\mathbf{4.8}$ & $\mathbf{5.8}$ && $33.5$ & $22.9$\\
    &$\%{>}15$ m & $\mathbf{1.2}$ & $\mathbf{3.1}$ && $94.2$ & $85.4$\\
    \multirow{2}{*}{Ours} & RMSE(m) & $11.0$ & $12.4$ && $\mathbf{13.2}$ & $\mathbf{12.4}$\\
    &$\%{>}15$ m & $23.4$ & $26.6$ && $\mathbf{33.1}$ & $\mathbf{28.7}$\\
    \hline
    \end{tabular}
    \caption{\ghl{Comparison of localization performance on the simulated dataset. The dataset consists of varying total number of measurements and maximum number of measurement faults (total \# of measurements, max \# of faults). Our approach demonstrates better localization performance in scenarios with large number of faults in comparison to J-PF and KF-RAIM.}}
    \label{tab:metrics}
\end{table}
In this experiment, we vary the number of available and faulty GNSS measurements at a time instant. We compare the localization performance of our algorithm with that of J-PF and SR-RAIM approaches. We consider the following scenarios (total \# of measurements, max \# of faults): $(5, 1)$, $(5, 2)$, $(7, 4)$, $(10, 6)$. For each of these scenarios, we induce a \ghl{bias error} of $100$ m with a standard deviation of $5$ m in the GNSS measurements. \ghl{We compute the localization metrics} for $50$ full runs of duration $400$ s each across different randomly generated trajectories.  

\begin{figure}[t!]
    \centering
    \begin{subfigure}{\textwidth}
        \centering
        \includegraphics[width=\linewidth]{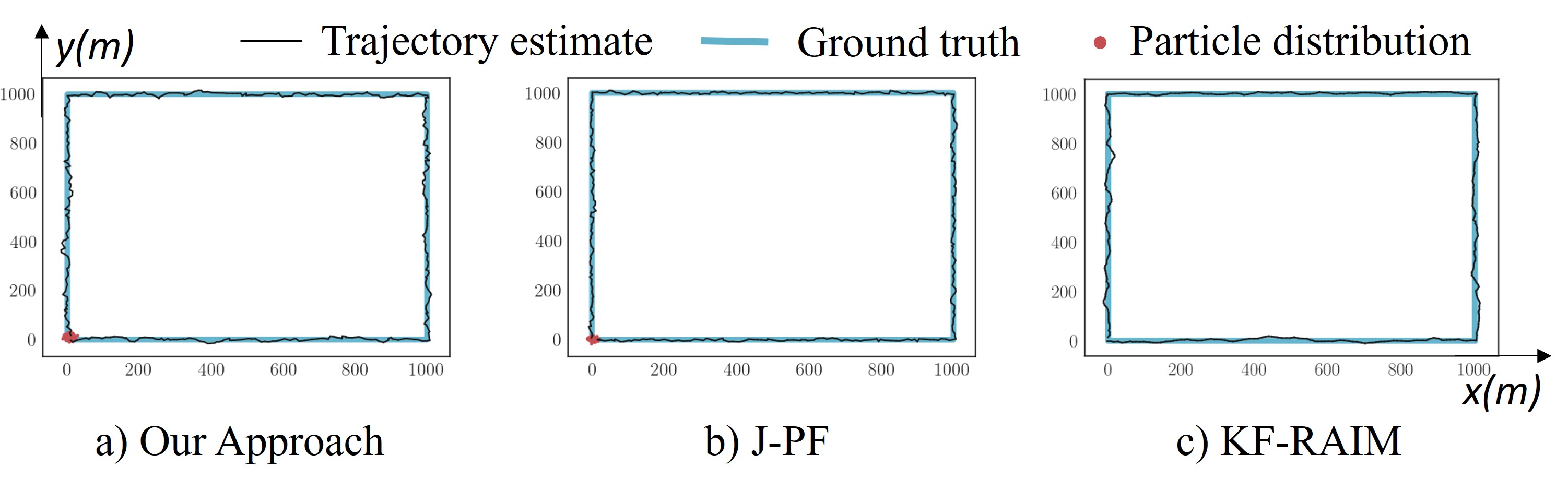}
        \caption{$1$ faulty GNSS measurement out of $5$ in total.}
    \end{subfigure}
    \begin{subfigure}{\textwidth}
        \centering
        \includegraphics[width=\linewidth]{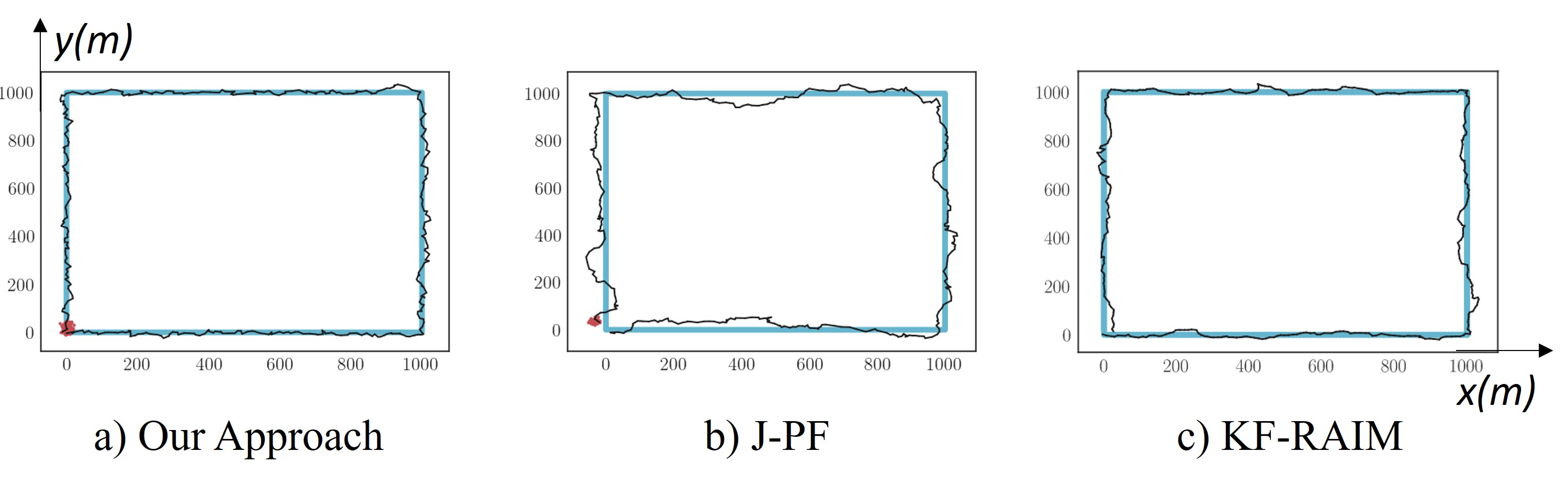}
        \caption{$6$ faulty GNSS measurements out of $10$ in total.}
    \end{subfigure}
    \caption{\ghl{Localization accuracy comparison between our approach, J-PF, and KF-RAIM approaches for (a) few-fault scenario and (b) many-fault scenario. We fix the underlying ground truth trajectory to be a square of side $1000$ m with $(0, 0)$ as both the start and end positions. Our approach estimates the vehicle state with better accuracy that J-PF and KF-RAIM in scenarios with a many faulty measurements.}}
    \label{fig:qual_f}
\end{figure}

The trajectories for two scenarios, $(5, 1)$ and $(10, 6)$, are visualized for each algorithm in Fig.~\ref{fig:qual_f} and the computed metrics are recorded in Table~\ref{tab:metrics}. As can be seen from the qualitative and quantitative results, our approach demonstrates better localization performance than the compared approaches for scenarios with high noise degradation while maintaining a low rate of large positioning errors. \ghl{J-PF has better performance than its counterparts in scenarios with single faults since it considers all the possibilities of faults separately. However, its performance worsens in scenarios with more than 2 measurement faults. Similarly, KF-RAIM is able to identify and remove faults in scenarios with few faulty measurements resulting in lower positioning errors, but has poor performance in many-fault scenarios.}

\begin{figure}[t!]
    \centering
    \includegraphics[width=0.9\linewidth]{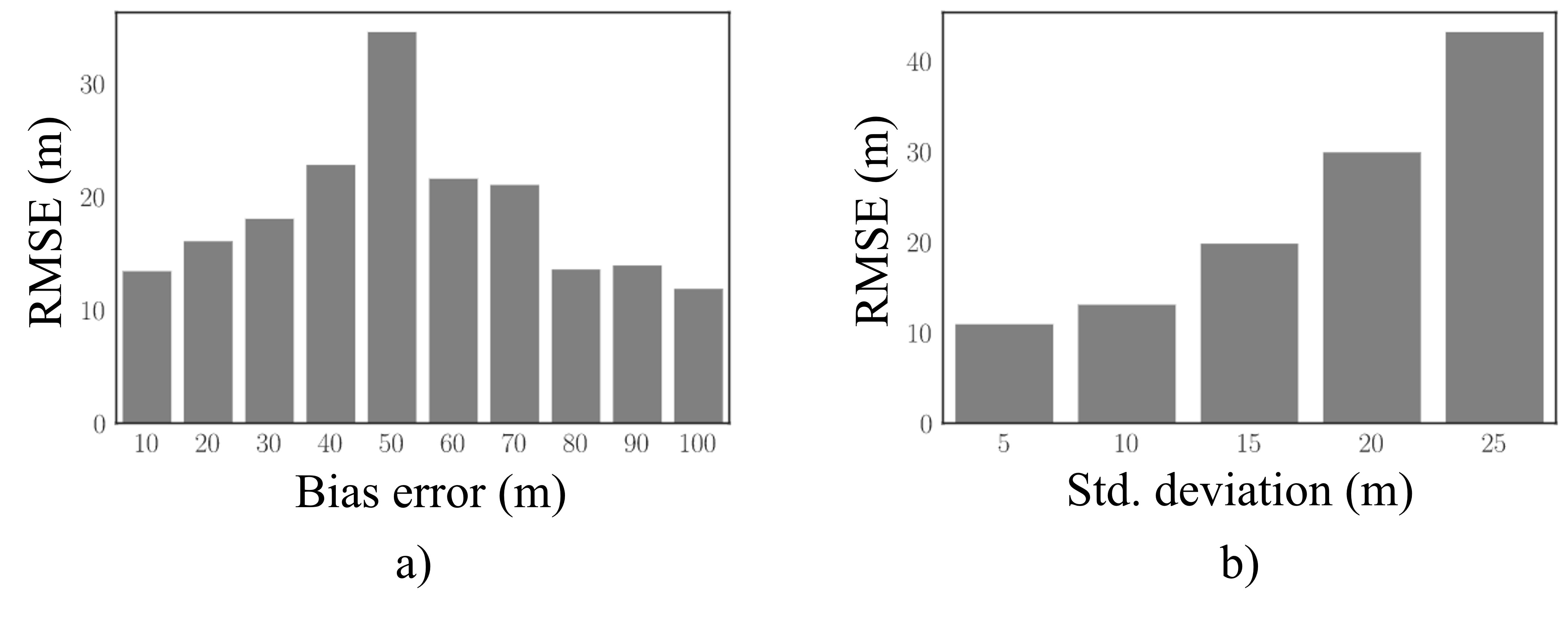}
    \caption{Sensitivity analysis of the localization performance of our approach with varying values of GNSS measurements (a) bias and (b) standard deviation. For low values of bias, the faults do not significantly impact the navigation solution, resulting in low RMSE values. Beyond the value of $50$ m, our approach is able to remove the impact of faulty measurements resulting in low RMSE values. The performance of the algorithm increasingly deteriorates with higher standard deviation in GNSS measurement noise since both the capability to remove faulty measurements as well as localization are hampered by an increased noise.}
    \label{fig:bias_var}
\end{figure}

\paragraph{Varying GNSS measurement noise}
In this experiment, we study the impact of GNSS measurement noise bias and standard deviation values on the performance of our algorithm. First, keeping the GNSS measurement standard deviation fixed at $5$ m, we vary the bias between $10-100$ m in increments of $10$ m. Next, keeping the bias fixed at $100$ m, we vary the standard deviation between $5-25$ m in increments of $5$ m. For all cases, the total number of measurements is kept at $7$ and the maximum number of faults is $3$. The RMSE for both studies are computed across $20$ different runs and plotted in Fig.~\ref{fig:bias_var}.  

The first plot shows that the localization performance of our approach initially deteriorates and then starts improving. This is because for low bias error values, the faulty GNSS measurements do not have a significant impact on the localization solution, even if the algorithm fails to remove them. Beyond the bias error value of $50$ m, the algorithm successfully assigns lower measurement weights to faulty measurements, resulting in lower RMSE for higher bias error values.  

The second plot demonstrates the impact of high measurement noise standard deviation on our approach. As the measurement noise standard deviation is increased, the RMSE increases in a superlinear fashion. The performance deteriorates with increasing noise standard deviation since both the tasks of fault mitigation as well as localization are negatively impacted by an increased measurement noise.           

\subsection{Evaluation on real-world data}
\begin{figure}[h!]
    \centering
    \includegraphics[width=\linewidth]{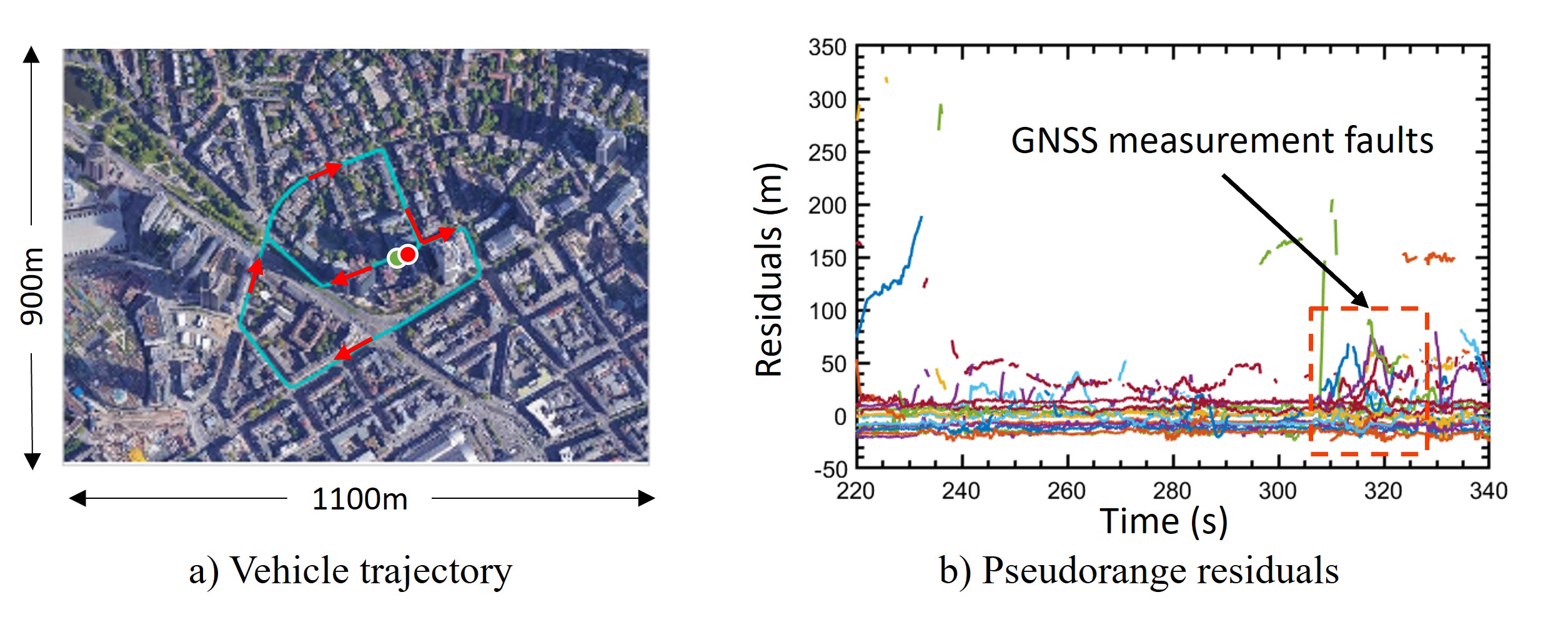}    
    \caption{Real-world dataset for validation of our approach. We use Frankfurt Westend-Tower sequence from smartLoc project dataset collected by TU Chemnitz~\cite{reisdorf_problem_2016}. The trajectory followed by the vehicle is shown in a). The vehicle moves from the start point (green) to end point (red) along the trajectory (cyan) in the direction specified by red arrows. b) shows the residuals of the pseudorange measurements with respect to ground truth position spread across 2 minutes (receiver clock drift errors removed via filtering). Multiple measurements demonstrate bias errors due to multipath and non line-of-sight effects in the urban environment.}
    \label{fig:real_data}
\end{figure}
For the real-world case, we use GNSS pseudorange and odometry measurements from the publicly available urban GNSS dataset collected by TU Chemnitz for the SmartLoc project~\cite{reisdorf_problem_2016}. \ghl{We consider the Frankfurt Westend-Tower trajectory from the dataset for evaluating our approach. The trajectory is ${\sim} 2300$ m long and has both dense and sparse urban regions, with $32\%$ of total measurements as NLOS signals.} 

\ghl{The dataset was collected on a vehicle moving in urban environments, using a low-cost u-blox EVK-M8T GNSS receiver at 5 Hz and CAN data for odometry at 50 Hz. The odometry data contains both velocity and yaw-rate measurements from the vehicle CAN bus.} The ground truth or reference position is provided using a high-cost NovAtel GNSS receiver at 20 Hz. Additionally, corrections to satellite ephemeris data are also provided as an SP3 file along with the data.  

\ghl{The receiver collects ${\sim} 12$ pseudorange measurements at each time instant from GPS, GLONASS, and Galileo constellations. As a preprocessing step, we remove inter-constellation clock error differences by subtracting the initial measurement residuals from all measurements, such that the measurements have zero residuals at the initial position.} Fig.~\ref{fig:real_data} shows the reference trajectory used for evaluation as well as measurement residuals with respect to the ground truth position in a dense urban region for a duration of $2$ minutes. 

\ghl{Our state-space consists of the vehicle position $(x, y)$, heading $\theta$, and clock bias error $\delta t$. To mitigate the impact of clock drift, we denote clock bias error $\delta t$ as the difference between the true receiver clock bias and a simple linear drift model with precomputed drift rate, and estimate this difference across time instants. We do not track the clock drift to keep the state size small for computational tractability. Note that in practice, the clock parameters can be determined separately using least-squares while the positioning parameters can be tracked by our approach for efficient computation. For our approach and J-PF, we use 1000 particles to account for a larger state space size than the simulated scenario. Additionally, we use 5 iterations of weighting in our approach, which is empirically determined to achieve small positioning errors on the scenario.} 
 
\begin{figure}[t!]
    \centering
    \includegraphics[width=\linewidth]{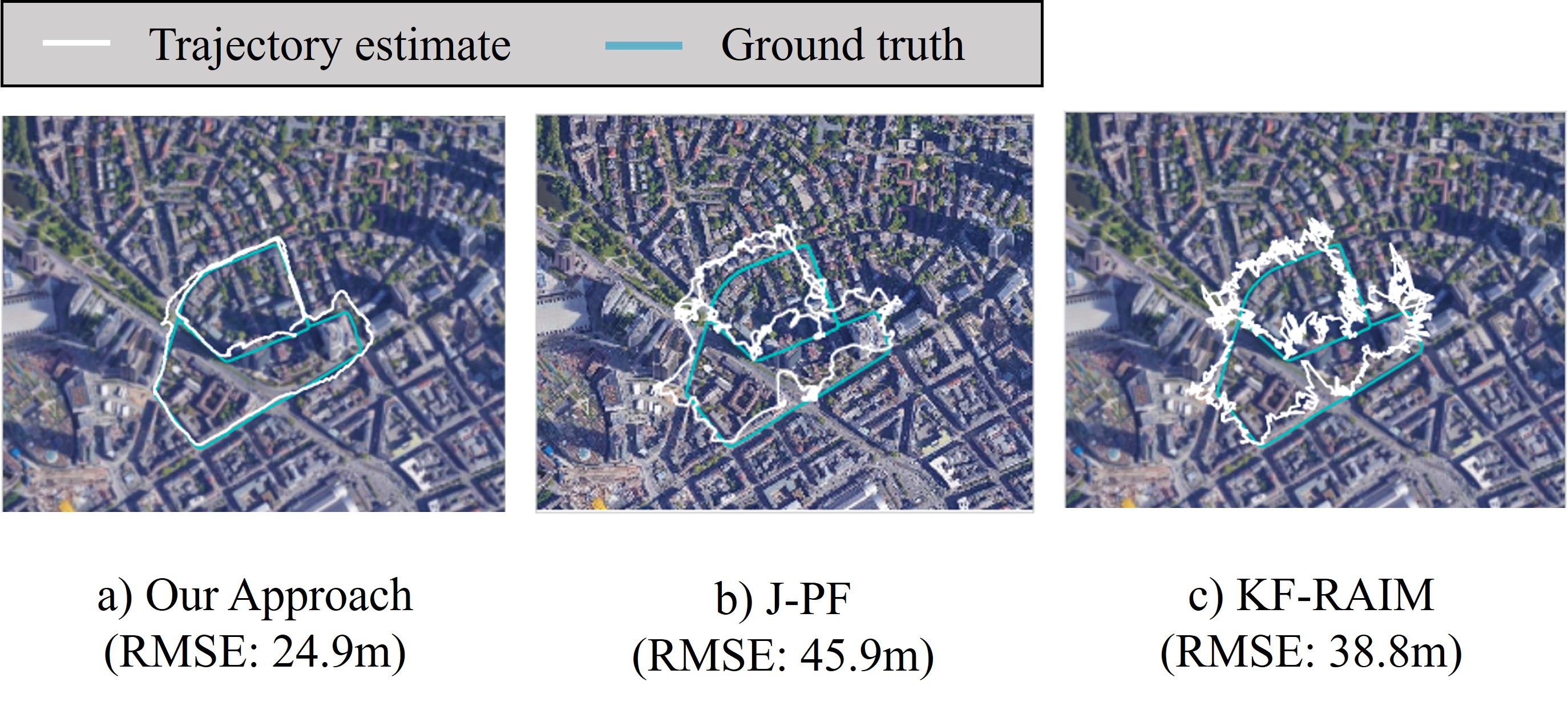}    
    \caption{\ghl{Estimated vehicle trajectories on real-world data. Our approach (a) produces trajectories that are closer to the ground truth as compared to trajectory from J-PF (b) and KF-RAIM (c). The horizontal RMSE values are computed by averaging over 20 runs with different randomness values in initialization and propagation. In regions where multiple measurements contaminated with bias errors, our approach is able to localize better than the baselines by assigning high measurement weights to measurements that are consistent with the particle filter probability distribution of vehicle state.} }
    \label{fig:traj_real}
\end{figure}

Fig.~\ref{fig:traj_real} shows \ghl{vehicle trajectories estimated using our approach, J-PF, and KF-RAIM in the local (North-East-Down) frame of reference.} Trajectories estimates from our approach are closer to the ground truth than compared approaches and thereby have lower horizontal RMSE values. \ghl{J-PF and KF-RAIM achieve similar RMSE values and visually exhibit a higher variance in state estimation error than our approach.}
 
\subsection{Comparison of integrity monitoring approaches}

\begin{figure}[t!]
    \centering
    \includegraphics[width=\linewidth]{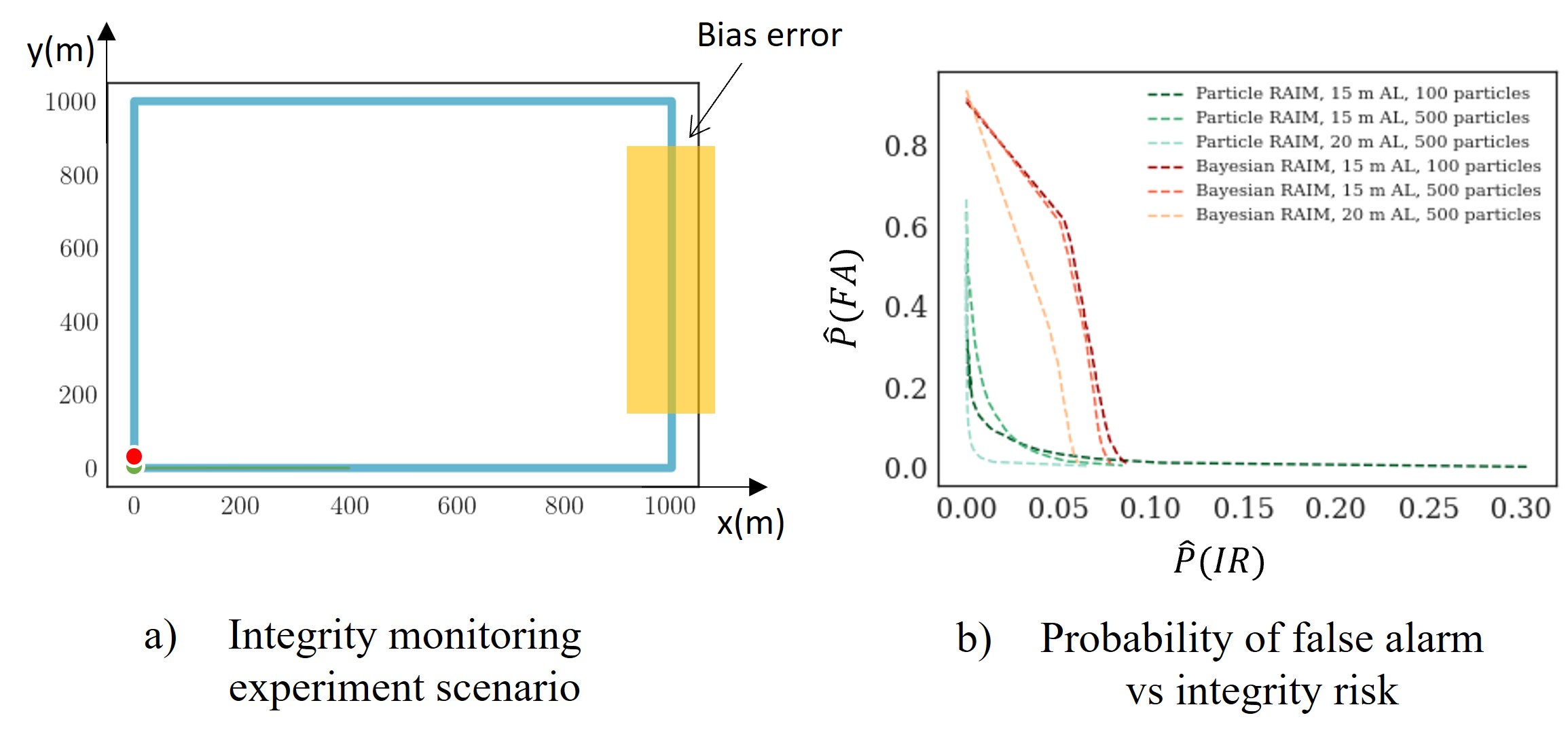}    
    \caption{\chl{(a) Simulation experiment for analyzing integrity monitoring performance. The start and end of the 400 s long trajectory are marked in green and red, respectively. Bias errors are induced in upto $60\%$ of the total 10 simulated GNSS measurements between $125-175$ s (yellow region). (b) Minimum attained probability of false alarm and integrity risk across different threshold values for our approach and Bayesian RAIM}~\cite{pesonen_framework_2011} \chl{for different number of particles ($100$, $500$) and alarm limit ($10$ m, $15$ m). The values are computed from more than $10^4$ samples across multiple runs. Our approach exhibits lower false alarms and smaller integrity risk than Bayesian RAIM across all thresholds for each configuration of the number of particles and alarm limit.}} 
    \label{fig:integ}
\end{figure}

\chl{Since obtaining a large amount of real-world data for statistical validation is difficult and out of scope for this work, we restrict our analysis of the integrity monitoring performance to simulated data. For the simulated scenario, we generate 400 s long trajectories with GNSS pseudorange measurements acquired at the rate of 1 Hz} (Fig.~\ref{fig:integ} a)). \chl{GNSS pseudorange measurements are simulated from 10 different satellites. At time instants between 125-175 s, we add bias errors in upto $60\%$ of the available measurements, such that the faulty measurements correspond to a position offset from the ground truth position. The position offset is randomly selected across different runs from between $50-150$ m. To limit our analysis to GNSS measurements, we do not simulate any odometry measurements for the simulations and localize only using GNSS pseudorange measurements for all the filters. For the particle filter algorithms, we set the propagation noise standard deviation to $20$ m to ensure that the position can be tracked in the absence of odometry measurements.}   
\chl{Both Bayesian RAIM~\cite{pesonen_framework_2011} and our approach depend on different thresholds for determining the system availability. In both approaches, a reference value of minimum accuracy and misleading information risk is required whose optimal value depends on the scenario and the algorithm. Varying these threshold values results in a trade-off between $\hat{P}(FA)$ and $\hat{P}(IR)$. For example, setting the threshold such that an alarm is always generated produces many false alarms, but no missed-identifications. Therefore, we compare the integrity monitor's performance by computing $\hat{P}(FA)$ and $\hat{P}(IR)$ across all different values of the thresholds and for two different settings for the number of particles ($100$, $500$) and the alarm limit ($10$ m, $15$ m).} (Fig.~\ref{fig:integ} b)). \chl{All the metrics are calculated using more than $10^4$ samples across multiple runs of the simulation for each algorithm. Fig.}~\ref{fig:integ} \chl{b) shows that our approach generates lower false alarms and missed-identifications than Bayesian RAIM across different threshold values for each considered parameter setting.}



\subsection{Discussion}
Our analysis and experiments on both the simulated and the real-world driving data suggest that our framework a) exhibits lower positioning errors than existing approaches in environments with a high fraction of faulty GNSS measurements, b) detects hazardous operating conditions with a superior performance to Bayesian RAIM and c) mitigates faults in multiple GNSS measurements in a tractable manner.

Although the localization performance of our approach in scenarios with multiple GNSS faults is promising, the performance in scenarios with few faults is worse than the existing approaches.
This poor performance results from conservative design choices in the framework design, such as the GMM measurement likelihood model instead of the commonly used multivariate Gaussian model. However, we believe that a conservative design is necessary for robust state estimation in challenging urban scenarios with multiple faults. Therefore, exploring hybrid approaches that switch between localization algorithms is an avenue for future research.

Another drawback of our framework is that the approach for determining system availability often generates false alarms. This is because of the large uncertainty within the GMM likelihood components, which in turn results in a conservative estimate of the misleading information risk. In future work, we will explore methods to reduce this uncertainty by incorporating additional measurements and sources of information such as camera images, carrier-phase, and road networks.

\section{Conclusion}
In this paper, we presented a novel probabilistic framework for fault-robust localization and integrity monitoring in challenging scenarios with faults in several GNSS measurements. The presented framework leverages GNSS and odometry measurements to compute a fault-robust probability distribution of the position and declares the navigation system unavailable if a reliable position cannot be estimated. \ghl{We employed a particle filter for state estimation and developed a novel GMM likelihood for computing particle weights from GNSS measurements while mitigating the impact of measurement errors due to multipath and NLOS signals. Our approach for mitigating these errors is based on the expectation-maximization algorithm and determines the GMM weight coefficients and particle weights in an iterative manner. To determine the system availability, we derived measures of the misleading information risk and accuracy that are compared with specified reference values.}
Through a series of experiments on challenging simulated and real-world urban driving scenarios, we have shown that our approach achieves lower positioning errors in state estimation as well as \ghl{small} probability of false alarm and integrity risk in integrity monitoring when compared to the existing particle filter-based approach. \ghl{Furthermore, our approach is capable of mitigating multiple measurement faults with lower computation requirements than the existing particle filter-based approaches. We believe that this work offers a promising direction for real-time deployment of algorithms in challenging urban environments.}

\section*{Acknowledgements}
This material is based upon work supported by the National Science Foundation under award \#2006162.


\ifCLASSOPTIONcaptionsoff
  \newpage
\fi



\bibliographystyle{IEEEtran}
\nocite{*}
\bibliography{bare_jrnl}
\end{document}